\documentclass[journal,twoside,web]{ieeecolor}
\usepackage{tmi}
\usepackage{cite}
\usepackage{amsmath,amssymb,amsfonts}
\usepackage{algorithmic}
\usepackage{graphicx}
\usepackage{textcomp}
\usepackage[hypertexnames=false]{hyperref}
\usepackage[ruled,vlined,linesnumbered]{algorithm2e}
\usepackage{booktabs}
\usepackage{tabularx}
\usepackage{multirow}
\usepackage{multirow}
\usepackage{booktabs}
\usepackage[table]{xcolor}
\usepackage{colortbl}
\usepackage{booktabs}
\usepackage{adjustbox}
\usepackage{ulem}
\usepackage{afterpage}
\usepackage{float}
\usepackage{dirtytalk}
\usepackage{longtable}
\newcommand{\mbluetext}[1]{\textcolor{mblue}{#1}}
\usepackage{soul} 
\soulregister{\cite}7 
\soulregister{\citep}7 
\soulregister{\citet}7 
\soulregister{\ref}7 
\soulregister{\pageref}7 
\soulregister{\say}7 
\soulregister{\mbluetext}7 
\hypersetup{
    colorlinks=true,
    linkcolor=mblue,
    filecolor=magenta,      
    urlcolor=cyan,
}
\usepackage[switch]{lineno}

\def\BibTeX{{\rm B\kern-.05em{\sc i\kern-.025em b}\kern-.08em
    T\kern-.1667em\lower.7ex\hbox{E}\kern-.125emX}}
\begin{document}

\title{Ultrasound Report Generation with Cross-Modality Feature Alignment via Unsupervised Guidance}
\author{Jun Li, Tongkun Su, Baoliang Zhao, Faqin Lv, Qiong Wang, \\Nassir Navab, \IEEEmembership{Fellow, IEEE}, Ying Hu, \IEEEmembership{Member, IEEE}, Zhongliang Jiang, \IEEEmembership{Member, IEEE},\\[-4.4ex]
\thanks{This work has been submitted to the IEEE for possible publication. Copyright may be transferred without notice, after which this version may no longer be accessible. This work was supported by the National Natural Science Foundation of China (No.62273328, No.U21A20489, No. U23A20345, U23A20391), the Regional Joint Fund of Guangdong (No. 2021B1515130003), the Key Fundamental Research Program of Shenzhen (No.JCYJ20220818101408019). This work is also supported by CAS Key Laboratory of Human-Machine Intelligence-Synergy Systems, Shenzhen Institutes of Advanced Technology. \textit{(Corresponding authors: Baoliang Zhao and Ying Hu)}}
\thanks{The first two authors contributed equally to this work.}
\thanks{Jun Li is with the Technical University of Munich and also the Munich Center for Machine Learning, Germany. (e-mail: june.li@tum.de).}
\thanks{Tongkun Su, Baoliang Zhao, Qiong Wang and Ying Hu are with Shenzhen Institute of Advanced Technology, the Chinese Academy of Sciences, China (e-mail: tk.su@siat.ac.cn, bl.zhao@siat.ac.cn, wangqiong@siat.ac.cn, ying.hu@siat.ac.cn).}
\thanks{Zhongliang Jiang and Nassir Navab are with the Chair for Computer Aided Medical Procedures and Augmented Reality (CAMP), Technical University of Munich, Germany (zl.jiang@tum.de, nassir.navab@tum.de).}
\thanks{Faqin Lv is with the Department of Ultrasound, The Third Medical Centre of Chinese PLA General Hospital and also with The Second School of Clinical Medicine, Southern Medical University, China (lvjin8912@163.com).}
}
\maketitle

\begin{abstract}
Automatic report generation has arisen as a significant research area in computer-aided diagnosis, aiming to alleviate the burden on clinicians by generating reports automatically based on medical images. In this work, we propose a novel framework for automatic ultrasound report generation, leveraging a combination of unsupervised and supervised learning methods to aid the report generation process. Our framework incorporates unsupervised learning methods to extract potential knowledge from ultrasound text reports, serving as the prior information to guide the model in aligning visual and textual features, thereby addressing the challenge of feature discrepancy. Additionally, we design a global semantic comparison mechanism to enhance the performance of generating more comprehensive and accurate medical reports. To enable the implementation of ultrasound report generation, we constructed three large-scale ultrasound image-text datasets from different organs for training and validation purposes. Extensive evaluations with other state-of-the-art approaches exhibit its superior performance across all three datasets. Code and dataset are valuable at this \href{https://lijunrio.github.io/Ultrasound-Report-Generation/}{link}.

\end{abstract}


\section{Introduction}
\label{sec:introduction}
\IEEEPARstart{M}{EDICAL} imaging provides non-invasive and real-time visualization of internal organs, tissues, and structures, which plays a vital role in modern healthcare for diagnosis and finding potential diseases \cite{jiang2023robotic}. However, the process of interpreting and writing reports from medical images can be time-consuming, knowledge-intensive and human-dependent, creating a significant burden on clinicians. As the scale of medical imaging continues to expand, radiologists and sonographers are struggling to meet the increasing demands of patients leading to potential delays in diagnosis and treatment. To alleviate their pressure, the development of automated medical report generation algorithms to assist them in writing reports has become increasingly important.

The success of image captioning has laid a solid foundation in medical report generation, which inspired researchers to explore the possibility of using similar architectures to generate medical reports automatically. The dominant approaches for report generation are based on the encoder-decoder structure \cite{vinyals2015show} that utilizes Convolutional Neural Network (CNN)\cite{he2016deep} to extract visual features from medical images, followed by Recurrent Neural Network (RNN)\cite{graves2012long} to generate descriptive text based on the extracted features. 
However, due to their significant differences from natural images, medical images pose unique challenges in aligning visual and textual features. Unlike natural images, medical images often exhibit similar visual features, making it difficult for non-experts to distinguish the subtle differences. Furthermore, medical reports tend to be longer and more detailed, describing complex observations of different physical tissues. As a result, there is a significant mismatch in feature diversity between image and text.

To address the performance degradation caused by this, researchers have explored various approaches to improve the performance of the encoder-decoder structure. Some methods involve adding annotated disease labels \cite{zhang2020radiology, yang2022knowledge} to assist the training process, while others \cite{yang_joint_2021, liu_exploring_2021} utilize the medical report subheadings as additional forms of image labels to better distinguish visual features. By incorporating this prior knowledge, the encoder-decoder structure can better capture the complex relationships between image and text, improving performance in aligning the visual and textual representations. While \cite{zhang2020radiology, liu_exploring_2021, yang_joint_2021, yang2022knowledge} these methods have shown promising results in report generation tasks, it's important to note that they require additional labelled data and may not be feasible for all types of datasets. The process of adding these annotations can impose an extra burden on clinicians.

\begin{figure}[!t]
\centerline{\includegraphics[width=\columnwidth]{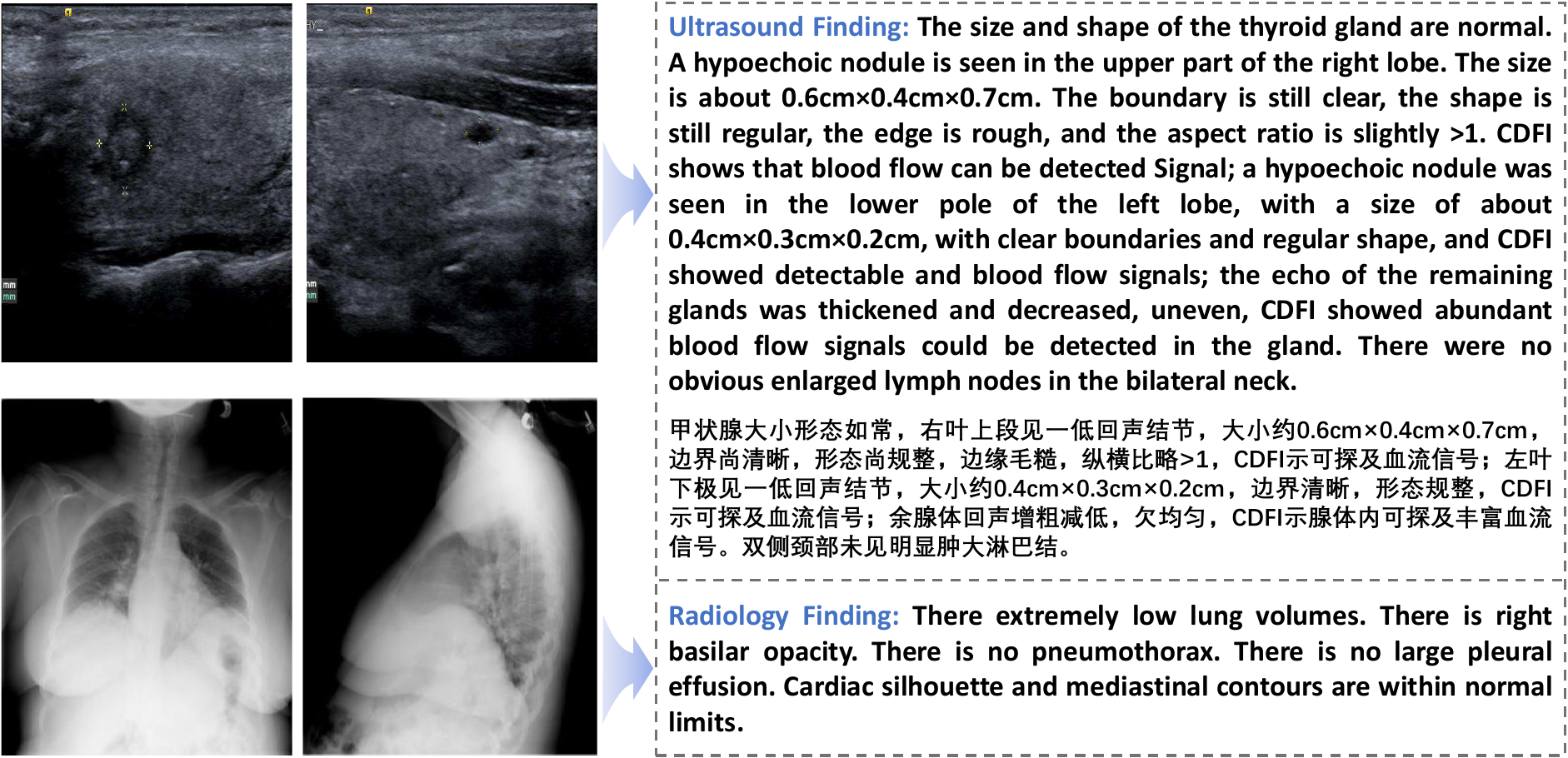}}
\caption{Examples of the ultrasound report and radiology report. The original ultrasound report is written in Chinese.}
\label{fig1}
\end{figure}

Furthermore, most of the existing works \cite{wang2021self, zhang2020radiology, liu_exploring_2021, yang_joint_2021, yang2022knowledge, liu_medical-vlbert_2021} in medical report generation focus on radiology reports, primarily attributed to the availability of well-known public datasets such as IU-Xray \cite{demner2016preparing} and MIMIC-CXR \cite{johnson2019mimic}. In contrast, the studies of ultrasound report generation have been relatively limited, despite ultrasound serving as a more extensively utilized and safer screening tool for diagnosing potential diseases. According to \mbluetext{Fig.~\ref{fig1}}, we can see that ultrasound report generation is different from radiology report generation at both image and text levels. Ultrasound images exhibit distinct characteristics such as low contrast and the presence of artefacts, which pose challenges in accurately extracting relevant visual features for textual description. Conversely, ultrasound reports tend to be lengthier and more detailed compared to radiology reports, often containing thorough descriptions of organs, lesions, and tissues, adding complexity to the text generation process. Moreover, current approaches in ultrasound tend to focus on description generation \cite{alsharid2022gaze}, which is similar to image captioning that aims to predict a short caption for education purposes. Therefore, there is a pressing need for further research to develop effective strategies for ultrasound report generation that can overcome these challenges.

In this work, we present a novel report generation framework that combines unsupervised and supervised learning methods to align the visual and textual features. Our approach is motivated by the learning and writing process of doctors. We leverage unsupervised learning to extract potential knowledge from the textual reports, which is similar to the process of doctors acquiring knowledge from medical records. By extracting potential knowledge from the text, we can provide a guide for the visual extractor to learn visual features related to the text.  This approach helps bridge the gap between visual and textual modalities without any additional disease labels from experts, which makes it more accessible and efficient in most datasets. In order to enhance the model’s ability to learn the global semantics of long and complex medical reports, we design a similarity comparison mechanism to aid the model in generating more accurate and longer reports. Our method calculates the overall similarity between the predicted reports and the ground truth reports in the training process to capture both the global semantics of the text report, resulting in a more accurate and comprehensive output that closely aligns with the ground truth. Besides, to demonstrate the effectiveness of our proposed method, we have built three separate ultrasound datasets, with each dataset specifically targeting different organs, including the breast, thyroid, and liver, respectively. The data collection has been approved by the institutional review board under YSB-2020-Y0902. In conclusion, our main contributions are summarized as:
\begin{itemize}
    \item We propose a novel framework that leverages both unsupervised and supervised learning methods to extract potential medical knowledge from text reports without requiring extra disease labels. This method is designed to align visual and textual features, thus alleviating visual and textual gaps in the medical report generation process. 
    \item Our framework generates long and accurate reports by employing a similarity comparison mechanism. This approach combines global semantic information to produce complex sentences, resulting in highly informative and accurate reports compared to other methods.
    \item We have collected three large separate ultrasound image text datasets, covering breast, thyroid, and liver. Specifically, the breast dataset includes 3521 patients, the thyroid dataset includes 2474 patients and the liver dataset includes 1395 patients. To the best of our knowledge, our research represents the first work to be evaluated and tested on multi-organ ultrasound report datasets.
\end{itemize}

This work is a significant extension of our previous conference paper \cite{li2022self} and offers several key contributions. First, we optimized each step in the Knowledge Distiller within our framework to better suit the task of ultrasound report generation, resulting in highly competitive results. Secondly,  we validated our method on three large-scale ultrasound report datasets of different organs, showcasing its generalizability. Thirdly, we conducted a comprehensive comparison with the current state-of-the-art methods in each dataset, showing the superior performance of our framework. Lastly, we conducted a thorough discussion of our experimental results, highlighting the strengths and limitations of our proposed method. 

\section{Related Work}
\subsection{Image Captioning}
Image captioning aims to generate brief descriptive sentences based on the image. Existing approaches in image captioning can be categorized into two main types: template/retrieval-based method and generative-based method. The template-based or retrieval-based method \cite{farhadi2010every, hodosh2013framing, girshick2015fast} involves detecting entities, attributes, and relationships from images using object detection models and then generating text sentences through template filling or retrieval from a database based on the identified relationships. Currently, the mainstream image captioning methods are based on the generative-based model \cite{xu2015show}, which utilises an encoder-decoder architecture as the backbone. This approach extracts visual features from the image using a visual encoder and generates descriptive sentences using a decoder based on these visual features. However, the performance of the basic encoder-decoder structure is often insufficient. Consequently, researchers have made various improvements, such as enhancing the encoder \cite{yao2018exploring} or decoder components \cite{ke2019reflective} of the network. Moreover, research in image captioning has also explored specialized tasks, including endowing models with human-like control over descriptions \cite{chen2021human} and accurately describing the time and numbers depicted in the image \cite{xu2021towards}. However, many of these methods involve recognition tasks and require additional image labels and detection boxes for auxiliary training.

\subsection{Radiology Report Generation}
Report generation for radiology images has been the major branch in the field of medical report generation, primarily due to the availability of a wide range of radiology datasets. Most existing methods in this area adopt the generative-based model \cite{xu2015show} employed in image captioning. However, directly transferring these methods to radiology report generation often fails to achieve comparable results. This difference arises from the inherent distinction between radiology images and natural images, as well as the disparity in the length of radiology reports compared to image captions. Thus, researchers have proposed various improvements to address these challenges. For instance, Jing \textit{et al.} \cite{jing2018automatic} employed a CNN to classify features extracted from radiology images, promoting the model to discriminate disease types. Zhang \textit{et al.} \cite{zhang2020radiology} constructed a graphical model of lung diseases to assist the decoder in generating long and accurate reports. this graph model has also been used as prior knowledge in the framework to enhance model generation in Liu's work \cite{liu_exploring_2021}. In another work, medical subject headings \cite{yang_joint_2021} were utilized as additional knowledge to facilitate the model in learning the relationship between images and text. Although these methods enhance the model's ability to generate radiology reports, they often require additional prior data, which needs separate collecting or manual annotating. Alternatively, some researchers have focused on improving the model structure to enhance the performance of the model.  Wang \textit{et al.} \cite{wang2021self} designed a model comprising two interrelated branches to improve training efficacy through a competitive approach. Li \textit{et al.} \cite{li2019knowledge} designed a retrieval policy module based on reinforcement learning to assist in model training. Chen \textit{et al.} \cite{chen2020generating} introduced a memory-driven unit to Transformer \cite{vaswani2017attention}, enabling the network to generate reports based on similar images. 

\subsection{Ultrasound Description Generation}
Differing from radiology report generation, research for ultrasound report generation is currently limited. Radiology reports mainly focus on pathological descriptions of the lung and heart, with a relatively narrow scope of diseases and organs. However, ultrasound can be utilized for different organs and tissues throughout the entire body. Consequently, reports for different organs may exhibit divergences in text style and format. Thus, radiology report generation and ultrasound report generation should not be recognized as identically the same tasks. Unlike X-rays, ultrasound imaging is naturally three-dimensional, providing two options for processing: treating it as three-dimensional videos or as two-dimensional images. Existing studies in video format focus on fetal screening. For instance, a CNN-LSTM-based ultrasound video captioning model \cite{alsharid2019captioning}  was proposed to simulate the doctor's oral description during second-trimester scans. Another study \cite{alsharid2022gaze} utilized doctor's gaze maps to guide the network to focus on regions of interest in the image, improving the quality of generated descriptions. In terms of two-dimensional images, a short disease description was generated by template-based method \cite{zeng2020deep}. However, these methods often require annotated labels and struggle to generalize to new datasets. Moreover, the generated sentences are notably short, resembling the image captions. Overall, research on generating long ultrasound reports is limited, and there is a scarcity of studies and evaluations across diverse ultrasound datasets involving multiple organs.

\begin{figure*}[!t]
\centerline{\includegraphics[width=0.95\textwidth]{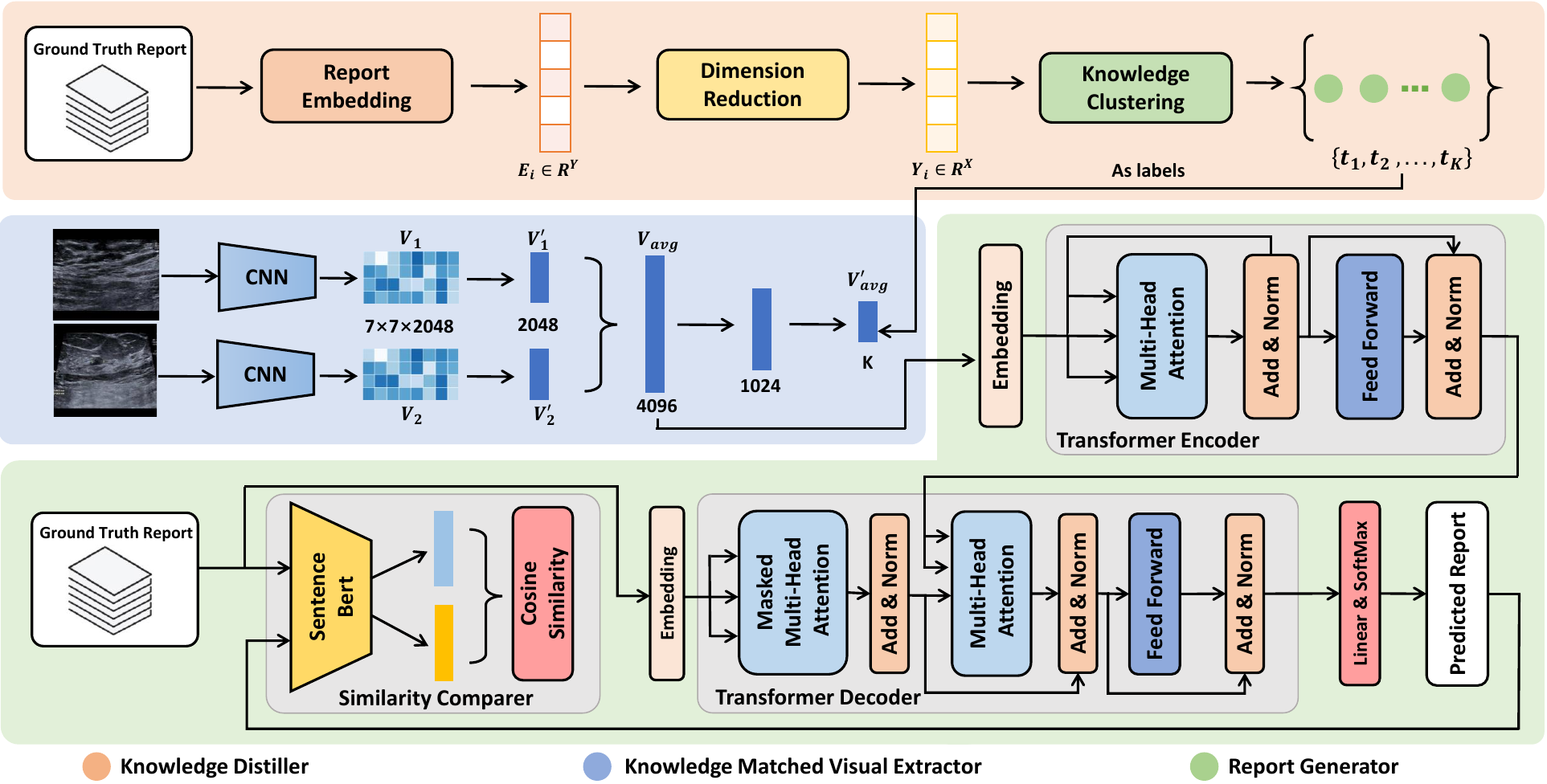}}
\caption{An overview of our proposed report generation framework. The orange section shows the Knowledge Distiller (KD), which extracts potential prior knowledge from ultrasound reports using unsupervised learning methods. The blue section is the Knowledge Matched Visual Extractor (KMVE), which uses prior knowledge extracted by the KD module to guide the visual extractor to capture knowledge-related visual features, addressing the problem of mismatch between visual and textual features. The green section shows the Report Generator (RG), which generates a text sequence from visual features, with a Transformer Encoder Decoder backbone and a proposed Similarity Comparer module.}
\label{fig2}
\end{figure*}

\section{Methodology}
\mbluetext{Fig.~\ref{fig2}} presents our proposed method consisting of three modules: Knowledge Distiller (KD), Knowledge Matched Visual Extractor (KMVE), and Report Generator (RG). KD aims to obtain prior knowledge from ultrasound reports. KMVE focuses on extracting visual features associated with text and aligning visual and textual features based on the acquired knowledge. RG is designed to generate ultrasound reports using aligned visual features with a comparison mechanism to enhance the generation performance.

\subsection{Obtaining Prior Knowledge from Ultrasound Reports}
Doctors gain proficiency by studying reports from experienced experts and summarizing their knowledge. To mimic this process, we design the KD model based on unsupervised clustering to extract the prior knowledge $T = \{t_1, t_2, \ldots, t_K\}$ from ultrasound reports $R=\{R_1, R_2, \ldots, R_n\}$ which consist of three stages: Report Embedding, Dimension Reduction, and Knowledge Clustering.

\subsubsection{Report Embedding}
\label{subsubsec:re}
Report embedding aims to transform the text report $R_i$ into the numerical feature $E_i \in \mathbb{R}^Y$. This is a crucial step in the overall KD pipeline
which can be represented as $\phi_{RE}(R_i)$, where $\phi_{RE}$ denotes the report embedding method.
Considering ultrasound reports are longer and more complex than radiology reports, to ensure the performance of the KD pipeline, we systematically evaluated three report embedding methods, including Bag of Words (BOW) \cite{zhang2010understanding}, Term Frequency-Inverse Document Frequency (TF-IDF) \cite{aizawa2003information}, and Sentence-Bert (S-Bert) \cite{reimers2019sentence}. BOW represents the report as a bag of constituent words, while TF-IDF calculates the importance of each word in the report based on its frequency in the document and the inverse frequency in the corpus. S-Bert utilizes pre-trained language models to embed reports into vector representations, which have been pre-trained on two large corpora \cite{bowman2015large, williams2018broad}. 

\subsubsection{Dimension Reduction}
Dimension reduction is vital to mitigate the computational complexity resulting from the high dimensional embedding vectors. In this work, we used the Uniform Manifold Approximation and Projection (UMAP) method \cite{mcinnes2018umap} to reduce the dimension of embedding vectors. UMAP is a nonlinear dimensionality reduction algorithm based on manifold learning, capable of reducing high-dimensional data to a lower-dimensional space while preserving the intrinsic data structure.
For a given embedding vector $E_i \in \mathbb{R}^Y$, we apply UMAP to reduce its dimension, resulting in $Y_i = \Phi_{umap}(E_i)$, where $Y_i \in \mathbb{R}^{X}$ and $X<Y$.

\subsubsection{Knowledge Clustering}
Knowledge clustering aims to extract potential prior knowledge from ultrasound reports by grouping similar text together. After reducing the dimension of the report embedding vectors, the clustering algorithm is applied to group them into $K$ clusters. Specifically, for the reduced vector set $Y=\{y_1, y_2, \ldots, y_{n}\}$, we utilize the K-Means \cite{hartigan1979algorithm} method to assign them to correspond clusters $t_k$ based on their similarity. This assignment is determined by minimizing the Euclidean distance between the vector $y_i$ and the centroid $m_j$ of cluster $j$: $t_k=\operatorname*{arg,min}_{j}\left|y_i - m_j\right|^2$. Here, $t_k$ represents the cluster assigned to the vector $y_i$, and $m_j$ is the centroid of cluster $j$. Following knowledge clustering, the text reports are organized into $K$ groups denoted as $T=\{t_1, t_2, \ldots, t_K\}$, where each group $t_i$ contains not only the writing style of doctors but also the potential knowledge within the reports. The details of parameter $K$ selection are shown in Section \ref{KD_experiments}.
In the knowledge clustering module, we adopt the K-Means clustering approach instead of the HDBSCAN method\cite{campello2013density} used in previous works\cite{li2022self}, 
as it offers lower computational complexity. For a fair comparison, we also evaluate other popular clustering methods\cite{dbscn, ng2001spectral, mullner2011modern} for the knowledge clustering process. Based on the evaluation results in Section \ref{section:clustering}, we demonstrate that the K-Means method is more suitable for our Chinese ultrasound datasets, offering competitive performance and lower computational costs.

\subsection{Extracting Knowledge Matched Visual Features}
To align the visual and textual representation, we propose the Knowledge Matched Visual Extractor (KMVE) module. This module utilizes the prior knowledge acquired by the knowledge distiller as pseudo-labels to promote the learning of visual features that are relevant to the knowledge and bridge the gap between visual and textual features.

Given the input image pairs $I = \{i_{m^1}, i_{m^2}\}$, where each image $i_m$ is represented by a tensor in $\mathbb{R}^{C\times H\times W}$, with $C$ denoting the number of channels, and $H$ and $W$ representing the height and width of the image, respectively.  The KMVE module begins by utilising a shared-weight CNN encoder to extract visual features from ultrasound images. Due to the challenges posed by low contrast and the presence of artefacts in ultrasound images, we choose the ResNet-101 model \cite{he2016deep}, pre-trained on ImageNet \cite{deng2009imagenet}, which has demonstrated excellent performance across various medical image analysis tasks, as the backbone network for feature extraction. Through this operation, the image pair is transformed into visual features $\{V_1, V_2\} \in \mathbb{R}^{7\times 7\times 2048}$. Then, a convolutional layer with a $7 \times 7$ kernel size average pooling is used to further process the features $\{V_1, V_2\}$, obtaining $\{V_1^{'}, V_2^{'}\}\in \mathbb{R}^{2048}$. These two features are then concatenated together to obtain the global average feature $V_{avg}\in \mathbb{R}^{4096}$. 
To align with the size of the knowledge topics $T$, $V_{avg}$ is further transformed into $V_{avg}'\in \mathbb{R}^{K}$. This reduction enables the KMVE module to calculate the loss function, which is defined as follows:
\begin{equation}
    \mathcal{L}_{kmve}=-\sum_{i=1}^k(t_i\times log(S_f(V_{avg}^{'}))) 
\end{equation}
where $t_i$ represents each cluster in the knowledge topic $T$ as a pseudo label. $S_f(\cdot)$ is the SoftMax  function. Due to the higher dimensionality of $V_{avg}$, it contains more comprehensive details from the visual features compared to $V_{avg}'$. Thus, the visual features $V_{avg}$ are chosen as the input for the report generator.

\subsection{Generating Reports from Visual Features}
After extracting visual features from ultrasound images, the generation of textual reports is the final step in our framework. We design a Report Generator (RG) that integrates a similarity comparison mechanism. The RG module considers both word-level and global semantic similarity to ensure consistent length and accuracy in the generated reports. The RG is built upon the transformer encoder-decoder architecture and the Similarity Comparer module (SC).

\subsubsection{Transformer Encoder-Decoder} Transformer (TF) \cite{vaswani2017attention} contains two main components: Transformer Encoder (TE) and Transformer Decoder (TD). In TE, the global visual feature, denoted as $V_{\text{avg}}$, is initially transformed into Query ($Q$), Key ($K$), and Value ($V$). Subsequently, Multi-Head Attention (MHA) is applied to compute the scaled dot-product attention between $Q$, $K$, and $V$. MHA consists of $n$ parallel heads, which can capture details from different subspaces. The results from all heads are then concatenated to obtain different spatial information. Following MHA, the output is passed through the Feed-Forward Network (FFN). Importantly, both MHA and FFN are followed by residual connection and Layer Normalization (LN) operations. In TD, the output of the TE is utilized as input for the decoder. Additionally, the current time step's input word embedding $x_t = w_t + p_t$  is also input into TD, where $w_t$ denotes the word embedding and $p_t$ denotes the position embedding. Similar to the TE module, MHA is applied to convert the input into the vector $h_m$. Next, the output of MHA was fed to FFN and LN, which can be represented as $h' = \text{LN}(h_m + \text{FFN}(h_m))$. Finally, the predicted word is generated using the formula $y_t \sim p_t = S_f(h^{'}W_p + b_p)$, where $W_p$ and $b_p$ are learnable parameters. To summarize, the TF loss is expressed as follows:
\begin{equation}
    \mathcal{L}_{T F}=-\sum_{i=1}^{n}\left(y_{i} \cdot \log \left(p_{i}\right)+\left(1-y_{i}\right) \cdot \log \left(1-p_{i}\right)\right)
\end{equation}

\subsubsection{Similarity Comparer (SC)} Ultrasound reports comprise detailed descriptions of various organs and tissues, often characterized by longer and more complex sentences. The comprehensive inclusion of all relevant descriptions in the generated report is crucial. However, the loss function in the TF focuses on the difference between separate words, lacking the ability to measure the overall semantic similarity between reports. To address this challenge, we design the Similarity Comparer (SC), which is able to compare the global semantics between the predicted report $p$ and the ground truth report $y$. By incorporating the SC module, our model can generate reports that offer a more comprehensive description.

In order to compute the similarity score, we utilised the S-Bert model to embed the predicted reports. Once embedded, the ground truth report and predicted report were represented as vectors $y_{e} \in \mathbb R^{768}$ and $p_{e} \in \mathbb{R}^{768}$, respectively. The cosine similarity between these vectors was then calculated to determine the similarity score, denoted as $S$. To ensure the similarity score is bounded between 0 and 1, we applied the RELU activation function. Specifically, the similarity score is computed as $S=f_{\text {relu }}\left(f_{c s}\left(y_{e}, p_{e}\right)\right)$, where $f_{\text {relu }}$ and $f_{c s}$ represent the RELU activation and cosine similarity functions, respectively. The loss function for the SC module is defined as the negative logarithm of the similarity score, summed over all sentences in the report. This can be represented as follows:
\begin{equation}
 \mathcal{L}_{S C}=-\sum_{i=1}^{N_{r}} \log \left(S_{i}\right)   
\end{equation}

\subsubsection{Training Strategy} 
\label{subsec:training_strategy}
In our framework, we combine the three losses mentioned above during the training stage. Algorithm 1 presents the training strategy of our method. The model first calculates $\mathcal L_{\text{KMVE}}$ and $\mathcal L_{\text{TF}}$ losses. Then, the network is frozen to stabilize its parameters for generating the full predicted report. Finally, the network is unfrozen to calculate the $\mathcal L_{\text{SC}}$ between the ground truth and the predicted report.
\begin{algorithm}
\caption{Training Strategy for our framework}
Initialize our framework ($M$)\;
Set the number of epochs $N$\;
Set the batch size $B$\;
\While{$\text{epoch} < N$}{
    Initialize the cumulative loss $\mathcal L_{\text{cum}} \gets 0$\;
    $\text{batch} \gets 0$\;
    \While{$\text{batch} < B$}{
        Calculate the KMVE loss $\mathcal L_{\text{KMVE}}$ \;
        Calculate the TF loss $\mathcal L_{\text{TF}}$ \;
        Freeze the weights of the model $M$\;
        Generate Predicted reports $R_{\text{pred}}$\;
        Unfreeze the weights of $M$ \;
        Calculate the SC loss $\mathcal L_{\text{SC}}=(R_{\text{pred}}, R_{\text{gt}})$ \;
        Compute the overall loss: $\mathcal L_{\text{cum}} = \lambda_1 \mathcal L_{\text{KMVE}} + \lambda_2 \mathcal L_{\text{TF}} + \lambda_3 \mathcal L_{\text{SC}} $ \;
        Optimize $M$ \;
        $\text{batch} \gets \text{batch} + 1$\;
    }
    Calculate and record the average loss $\bar{\mathcal L}_{\text{cum}} = \frac{\mathcal L_{\text{cum}}}{B}$ \;
    Save the model after this epoch \;
    $\text{epoch} \gets \text{epoch} + 1$ \;
}
\label{alg:example}
\end{algorithm}

\section{Experiments}
\subsection{Overview of the Datasets}
\begin{figure}
\centerline{\includegraphics[width=\columnwidth]{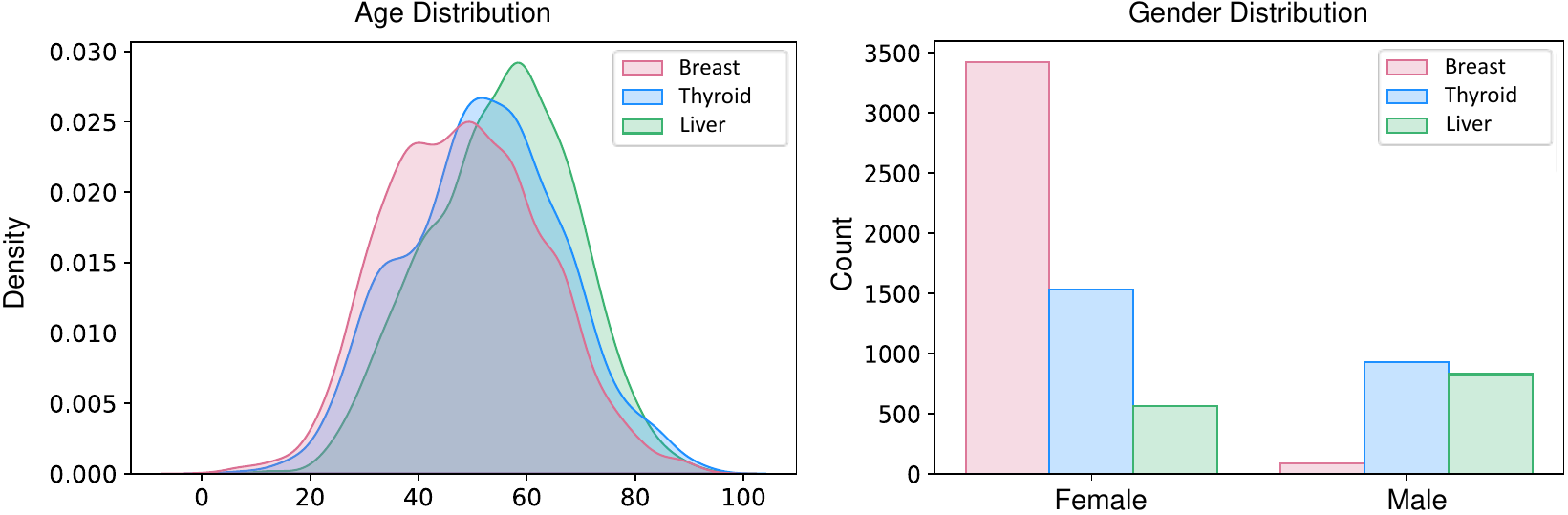}}
\caption{Age and gender distribution of our collected ultrasound datasets from three organs.}
\label{fig3}
\end{figure}

To evaluate the performance of the proposed framework on different types of ultrasound datasets, we collected three different datasets of the breast, thyroid and liver.  Specifically, the breast dataset consists of 3521 patients,  the thyroid dataset consists of 2474 patients, and the liver dataset consists of 1395 patients. All data used in this study were sourced from the ultrasonic department's database at the PLA General Hospital. The ultrasound image was saved in JPEG format, as illustrated in \mbluetext{Fig.~\ref{fig1}}. Further insights into the age and gender distribution within each dataset are provided in \mbluetext{Fig.~\ref{fig3}}. In the original data, each report is associated with a set of ultrasound images. We selected two images from the reports, as chosen by the doctors, to serve as the image pair associated with each report.

During the preprocessing, we conducted word segmentation on the ultrasound reports. Besides, we replaced numerical values such as lesion size and location in the text with special tokens, as shown in Table \ref{table:number_replace}. This decision was made due to the existing limitations regarding the accuracy of numerical predictions achieved through generative models. Although GPT \cite{brown2020language} exhibits a commendable level of precision in certain mathematical tasks, its inference capabilities heavily rely on extensive training with large datasets, which proves challenging in the medical domain due to the limited dataset scale. Therefore, our framework focuses solely on generating the textual descriptions of the reports. Besides, we inserted start \texttt{\textless start\textgreater} and end \texttt{\textless end\textgreater} tokens at the beginning and end of each report. Finally, each dataset was divided into training, validation, and test sets in a ratio of 7:1:2. Notably, we ensured that there was no overlapping of data between these sets, guaranteeing the reliability of the training results. 

\begin{table}
    \caption{Replacement rule for numerical values}
    \centering
    \renewcommand{\arraystretch}{1.1}
    \begin{tabular}{cc}
    \hline
    \textbf{Numerical Valu}e & \textbf{Replacement Token} \\
    \hline
        $1.5cm \times 0.6 cm$ &  \texttt{\_2DS\_} \\
        $1.0cm\times 0.8cm\times 0.9cm$ & \texttt{\_3DS\_} \\
        $12$ o'clock position  & \texttt{\_Loc\_} \\
        $3.7cm$ & \texttt{\_SCM\_} \\
        $2.8mm$ & \texttt{\_SMM\_} \\
    \hline
    \end{tabular}
    \label{table:number_replace}
\end{table}

\subsection{Experimental Settings}
\subsubsection{Evaluation Metrics}
\label{subsec:metrics}
We assess the quality of predicted reports through three metrics: Natural Language Generation (NLG) metrics, Clinical Efficacy (CE) metrics, and entailment between predicted report and ground truth.

For the standard NLG metrics, we selected: BLEU\cite{papineni2002bleu}, ROUGE-L\cite{lin2004rouge}, and METEOR\cite{banerjee2005meteor}, which are widely adopted in most works. These selected metrics can comprehensively assess the similarity between the generated reports and the ground truth reports. BLEU is a commonly used metric for assessing word overlap between the generated and the ground-truth text. It measures the degree of overlap at different n-gram levels, including BLEU-1, BLEU-2, BLEU-3, and BLEU-4, thereby capturing various levels of linguistic similarity between the generated and the reference reports. ROUGE-L is a metric based on the longest common subsequence algorithm. It considers the similarity of sentence-level structures and identifies the longest co-occurring n-grams in sequences. This metric effectively captures the overall structural similarity between the generated and reference reports. METEOR evaluates the quality of the generated text by considering both precision and recall, with linguistic features such as word order and synonymy. All the metrics mentioned above have a value range from 0 to 1, where a higher value indicates better performance.

For the CE metrics, we aim to focus on the key information in the reports rather than the text similarity. We extracted essential entities for each report based on suggestions from sonographers (more details see Table \ref{table:comparison}). Each dataset contains a set of m key entities of interest, denoted as $\{1,2,3,…,m\}$. If an entity $i$ is mentioned in the report, it is labelled as 1 $(y_i=1)$; otherwise, it is labelled as 0 $(y_i=0)$. This setup allows the task to be converted to multi-label classification. Finally, we calculate accuracy, precision, recall, and F1 score.

In addition to NLG and CE metrics, we also utilized the Natural Language Inference (NLI) model to determine whether the predicted report logically follows from the ground truth. In the medical domain, accurately describing each pathology is crucial. For instance, terms like \say{high echogenicity} and \say{low echogenicity} both pertain to \say{echogenicity}, yet their interpretations are diametrically opposite. We aggregate sentences from each entity and utilize DeBERTa \cite{he2021deberta}, a widely-used BERT-based model for NLI, to compare these aggregated sentences with the related aggregated ground truth sentences.

\subsubsection{Implementation Details}
\label{sec:implementaion}
Our model is implemented using the PyTorch framework and trained on two NVIDIA GeForce RTX 3090 GPUs. To optimize the KD module, we separately conduct experiments on three different datasets to determine the best choices for embedding method, dimension reduction, and the number of clusters. These optimized results serve as prior knowledge for the framework, and more details can be found in Section \ref{KD_experiments}. For the RG model, the number of layers in both the TE and TD is set to 3. We set the feature dimension of the MHA to 512 and used 8 heads. The maximum number of training epochs for the entire network is set to 50, and training stops when the validation loss does not decrease within 10 epochs. The batch size during the training process was set to 128, and the maximum sentence length for sentence generation was set to 150. To optimize the models, we utilize the ADAM optimizer \cite{kingma2014adam} with a learning rate of 5e-4 for KMVE and 1e-4 for RG. During training, the learning rate is decayed by a factor of 0.8 after each epoch. 

\begin{figure}[t]
\centering
\setlength{\abovecaptionskip}{-4pt}
\includegraphics[width=\columnwidth]{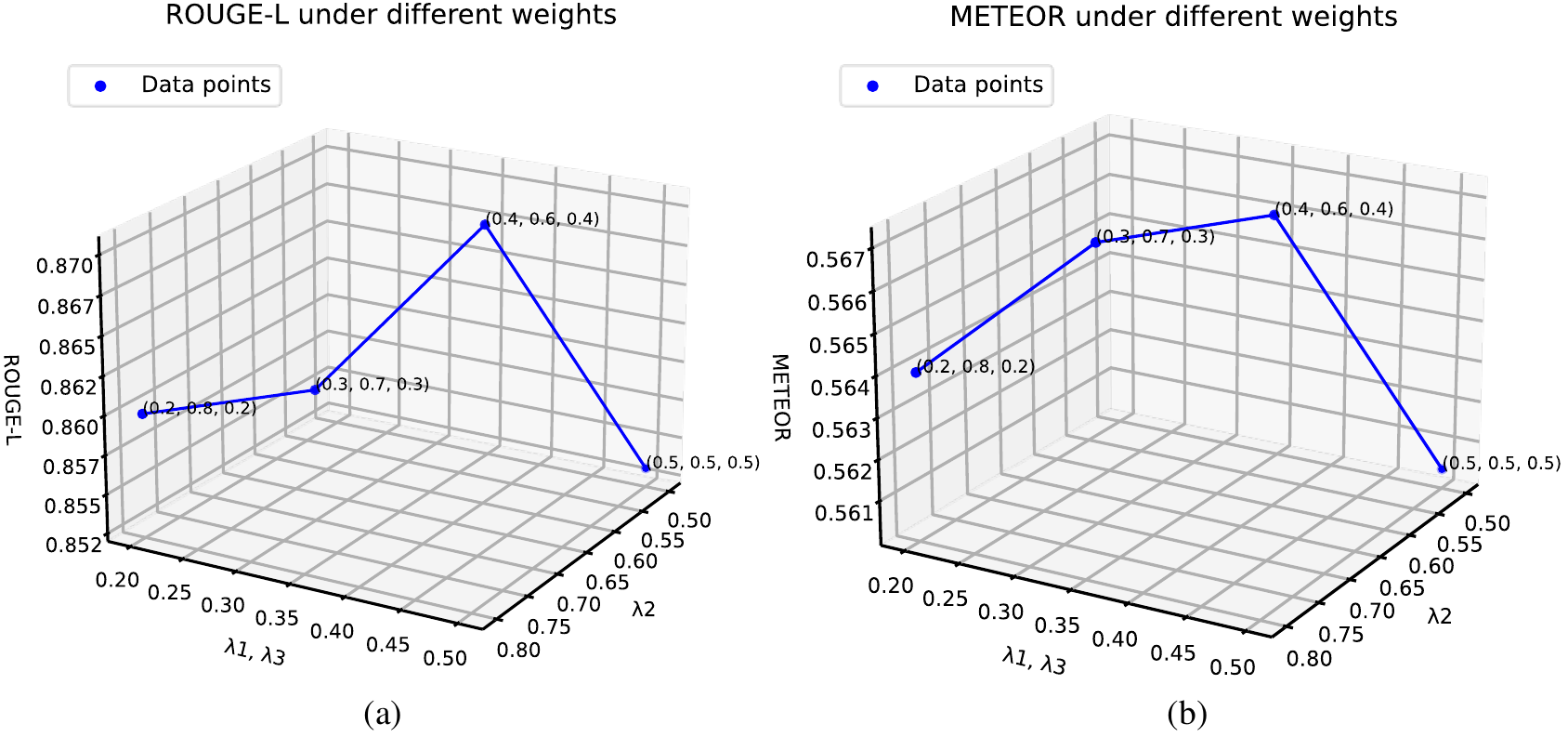} 
\caption{Hyper-parameter searching with 10\% liver training data. (a) shows the report generation performance evaluated by the ROUGE-L metric, whereas (b) shows the results evaluated by the METEOR metric.}
\label{fig-parameter}
\end{figure}

The balancing weights $\lambda_1$, $\lambda_2$, and $\lambda_3$ are set to 0.4, 0.6, and 0.4 respectively. These weights were determined through parameter searching on 10\% of the liver dataset. In our work, the $\mathcal{L}_{KMVE}$ and $\mathcal{L}_{SC}$ are our proposed losses to support the report generation, $\mathcal{L}_{TF}$ is the fundamental loss for language modelling. We assume that the weight of $\mathcal{L}_{TF}$ should be relatively larger than $\mathcal{L}_{KMVE}$ and $\mathcal{L}_{SC}$ to maintain the effectiveness of the framework. Because at the beginning of the training, the model needs to first understand generating word by word, and then focus on the similarity between the entire report and the knowledge matched between images. To empirically find the optimal values for $\lambda_1$, $\lambda_2$, and $\lambda_3$, we initially assigned equal weights (0.5, 0.5, 0.5) to all and began training on a randomly selected 10\% sample of the liver dataset. During this process, we progressively increased $\lambda_2$ while simultaneously reducing $\lambda_1$ and $\lambda_3$ to balance the overall increase in the entire loss value. Here, we set $\lambda_1 = \lambda_3$, because we consider that both these two losses have equal contributions to the report generation. \mbluetext{Fig.} \ref{fig-parameter} shows that the combination (0.4, 0.6, 0.4) achieves the highest scores on both METEOR and ROUGE-L. Thus, We use the (0.4, 0.6, 0.4) as the weights of each loss.

\subsection{Experiments for the Knowledge Distiller}
\label{KD_experiments}
The KD involves the selection of the embedding method, dimension reduction, and the number of clusters to achieve the best clustering results. To determine the optimal parameters for each stage, we followed a two-step process.

\begin{table}
    \caption{The coarse range for the number of clusters from different embedding methods. }
    \centering
    \renewcommand{\arraystretch}{1.1}
    \begin{tabular}{ccccc}
    \hline
    \textbf{Dataset} & \textbf{Method} & \textbf{Silhouette} & \textbf{Elbow} & \textbf{Range} \\
    \hline
             &  BOW   & 2 & 18 & [2, 18] \\
      Breast & TF-IDF & 7 & 17 & [7, 17] \\
             & S-Bert & 4 & 18 & [4, 18] \\
    \hline
            &  BOW   & 2 & 16 & [2, 16] \\
    Thyroid & TF-IDF & 15& 18 & [15, 18] \\
            & S-Bert & 2 & 18 & [2, 18] \\
    \hline
            &  BOW   & 2 & 18 & [2, 18] \\
    Liver   & TF-IDF & 12 & 18 & [12, 18] \\
           & S-Bert & 3 & 14 & [3, 14] \\
    \hline
  \end{tabular}
  \label{table:coarse}
\end{table}

In the first step, we used two widely employed clustering evaluation methods to determine the coarse range of cluster numbers from different embedding methods. This process helped narrow down the options for subsequent analysis. In detail, the silhouette coefficient method (Silhouette) \cite{rousseeuw1987silhouettes} was utilized to calculate the lower bound, while the elbow method (Elbow) \cite{thorndike1953belongs} was applied to determine the upper bound of different embedding methods. We selected BOW, TF-IDF, and S-Bert as the report embedding methods to convert the ultrasound reports to embedding vectors. The experimental results for the first step are presented in Table \ref{table:coarse}.

\begin{figure}[t]
\centerline{\includegraphics[width=\columnwidth]{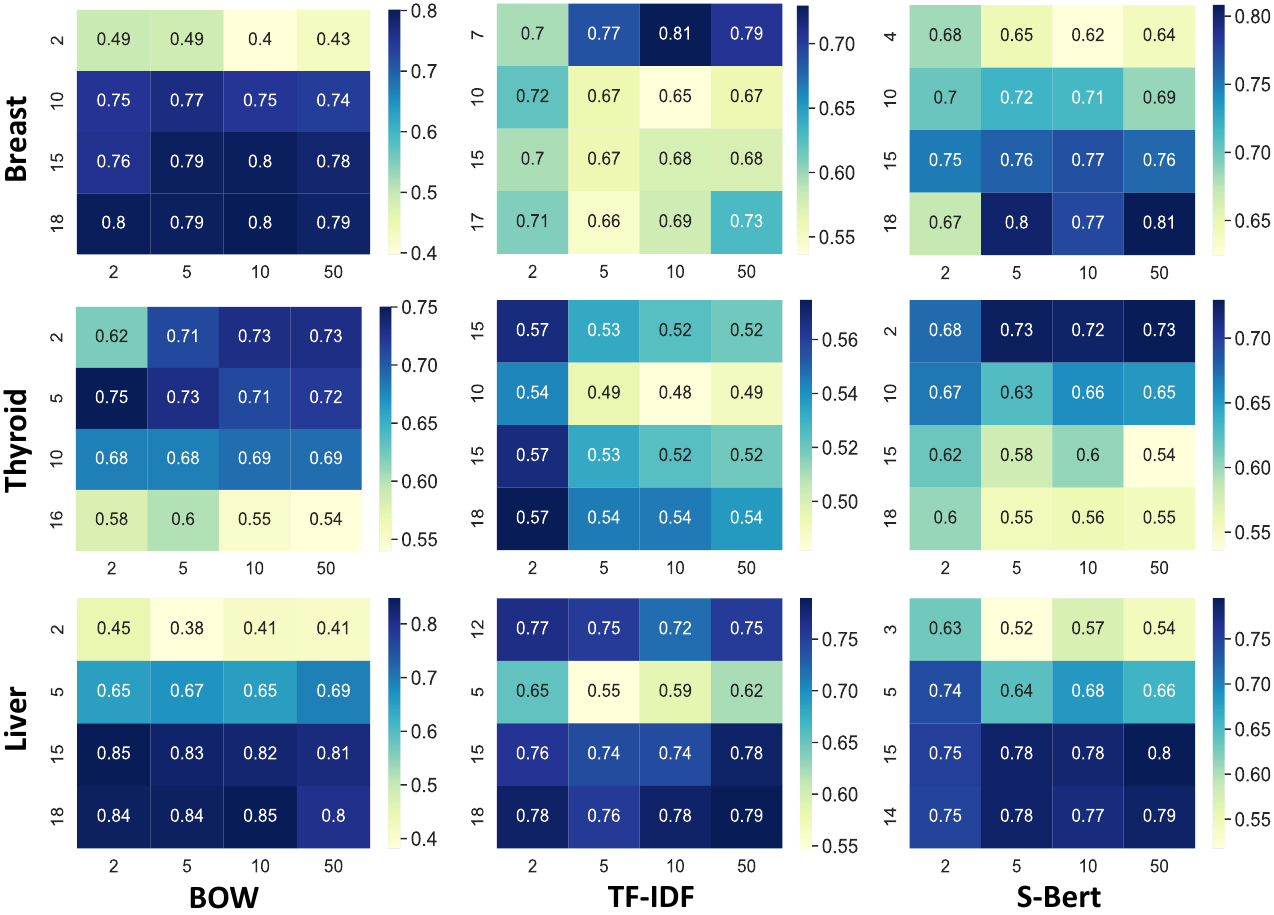}}
\caption{Heatmap of Clustering Results with different Dimensionality Reduction and Cluster Numbers. Each heatmap in this table displays clustering results, with the x-axis representing the dimensions of dimensionality reduction and the y-axis indicating different numbers of clusters. The values in each cell of the heatmap represent the silhouette coefficient scores, which reflect the performance of the clustering for each combination of dimension reduction and cluster numbers.}
\label{fig4}
\end{figure}

In the second step, the final clustering outcome is determined by selecting the result with the highest silhouette score. This involves evaluating the performance using four commonly employed dimensionality reduction dimensions: 2, 5, 10, and 50.  Additionally, for the selection of the number of clusters, we uniformly sample four cluster numbers from the initial coarse range obtained in the first stage.  As a result, for each embedding method, we obtain a total of 16 different clustering results, each corresponding to a distinct combination of dimension reduction and cluster numbers. Finally, we select the cluster with the highest score among the three embedding methods as the outcome for the final KD module.

\begin{table}[t]
  \caption{Final parameter settings and cluster scores for the knowledge distiller module. From left to right, the column's headings are Dataset, Dataset Size, Embedding Method, Vocabulary Size, Dimension Reduction, Clustering Number and Silhouette Score}
\centering
\renewcommand{\arraystretch}{0.5}
\begin{tabularx}{\columnwidth}{*{7}{>{\centering\arraybackslash}X}} 
\hline
\textbf{Dataset} & \textbf{Data. Size} & \textbf{Embedd. Method} & \textbf{Vocab. Size} & \textbf{Dimen. Reduct.} & \textbf{Cluster Num.} & \textbf{Silhoue. Score} \\
    \hline
     Breast  & 3521 & S-Bert & 694 & 50 & 18 & 0.81 \\ 
     Thyroid & 2474 & BOW & 659 & 2 & 5 & 0.75 \\   
     Liver   & 1395 & BOW & 470 & 10 & 18 & 0.85 \\  
    \hline
  \end{tabularx}
  \label{table:kd_res}
\end{table}

\mbluetext{Fig.~\ref{fig4}} illustrates the evaluation results obtained from three datasets. The top row shows the experimental outcomes for the breast dataset, followed by the second row shows the results for the thyroid dataset, and the third row shows the results for the liver dataset.  Within each row, the first column represents the results obtained from the BOW embedding method, the second column represents the results obtained from the TF-IDF method, and the third column represents the results obtained from the S-Bert method. Each heatmap in the figure provides insights into the clustering performance. The x-axis of each heatmap denotes dimensionality reduction dimensions, while the y-axis represents different numbers of clusters. The value displayed in each heatmap cell corresponds to the silhouette coefficient score of the clustering results by the selected combination of dimension reduction and the number of clusters. Based on \mbluetext{Fig.~\ref{fig4}}, the final parameter settings for the KD module on the three datasets are summarized in Table \ref{table:kd_res}. These parameter configurations yield the highest silhouette scores for each dataset. In situations where equivalent scores were obtained, our selection prioritized results with higher dimensions, as such dimensions tend to retain more comprehensive details of embedding. These selected outcomes subsequently serve as the prior knowledge for each dataset. 
According to Table \ref{table:kd_res}, it is clear that while S-Bert demonstrates the best performance on the breast dataset, the conventional BOW model is notably effective on both the thyroid and liver datasets. We hypothesize that this variation in performance may be attributed to differences in dataset size and textual characteristics. Specifically, the breast dataset is larger and contains more complex textual data, which may not be optimally handled by the simpler BOW model. Conversely, in the smaller and textually less diverse thyroid and liver datasets, the straightforward approach like the BOW model is not only adequate but potentially superior. This observation also suggests that the representations provided by S-Bert's pre-trained embeddings may not confer significant advantages in scenarios where the textual diversity is limited.

\subsection{Experiments on Different Clustering Methods}
\label{section:clustering}
\begin{figure}[ht]
\centering
\setlength{\abovecaptionskip}{-6pt}
\includegraphics[width=\columnwidth]{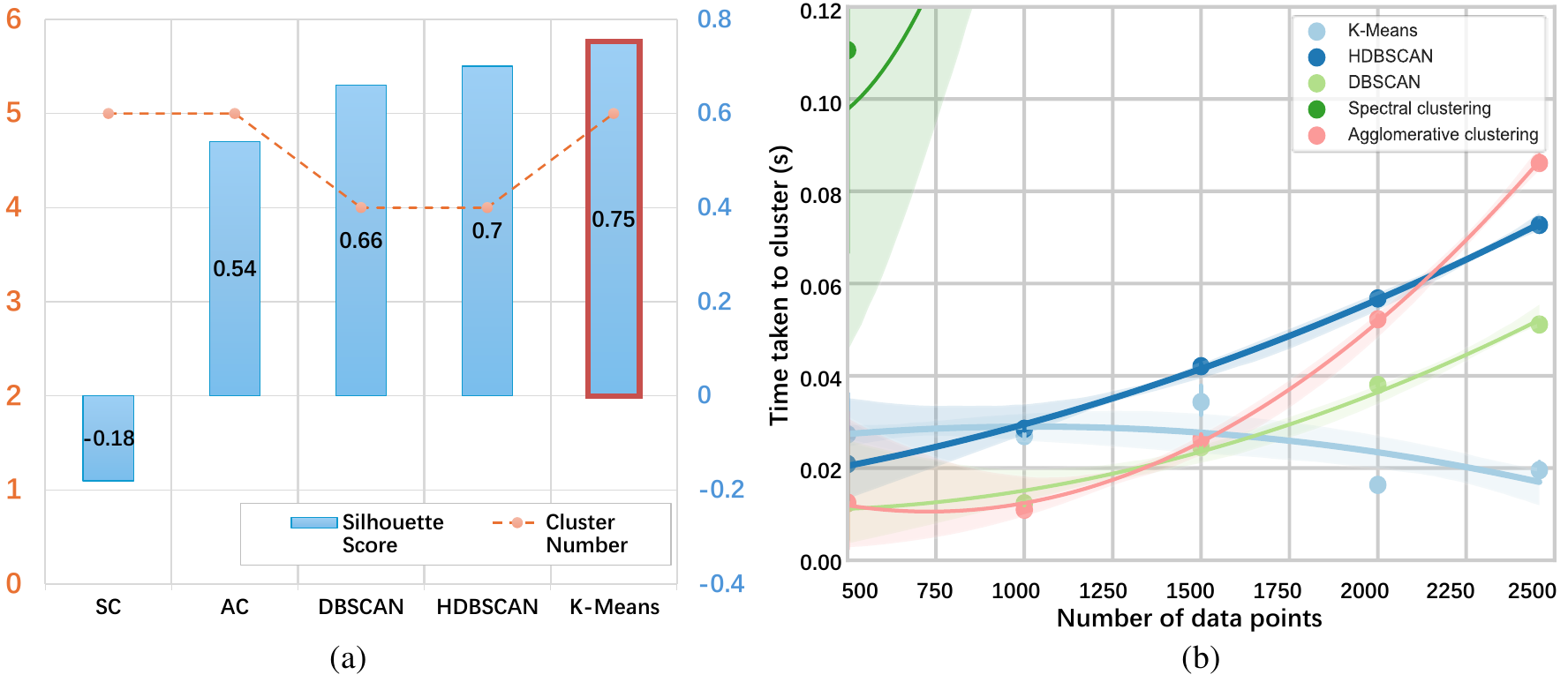} 
\caption{Comparison between different clustering methods. (a) Clustering performance of each method. (b) Time efficiency of each method. SC and AC denote spectral clustering and agglomerative clustering.}
\label{fig5-add}
\end{figure}

In the previous section, we conducted detailed experiments in each part of the KD pipeline. We used the K-Means algorithm as our knowledge clustering method, as it offers lower computational complexity and better performance in our dataset. To verify it, we compare the K-Means method with other popular clustering methods\cite{dbscn, ng2001spectral, mullner2011modern}. In \mbluetext{Fig.} \ref{fig5-add} (a), we evaluate the silhouette score of different clustering methods on our ultrasound dataset. To ensure a fair comparison, all settings are kept the same in the K-Means algorithm and tested on the thyroid dataset. However, due to DBSCAN and HDBSCAN's inability to directly set the cluster number, we ultimately keep the clustering results with 4 cluster numbers, which closely approximates 5. It can be observed that, compared to other methods, K-Means achieves a relatively higher silhouette score (0.75), while the second-best method is just 0.7. In \mbluetext{Fig.} \ref{fig5-add} (b), we aim to assess the computational time required by different methods as the dataset size increases. Notably, K-Means demonstrates relatively lower time consumption compared to other methods when dealing with more than 2000 data points.

Furthermore, we evaluate the influence of clustering results on final report generation based on the outcomes obtained in \mbluetext{Fig.} \ref{fig5-add} (a). From Table \ref{table:cluster}, it's evident that K-Means maintains competitive performance with the highest BLEU scores. Compared to our prior work\cite{li2022self}, which utilized HDBSCAN, K-Means proves better suited for the Chinese ultrasound dataset, exhibiting higher BLEU scores and ROUGE-L metrics. While some methods may excel in METEOR and ROUGE-L, K-Means remains preferable for larger datasets due to its lower computation. Notably, despite the varying impacts of different clustering methods on report generation, all methods surpass the baseline \say{TF+SC}, which lacks unsupervised clustering guidance (refer to Section \ref{subsec:ablation} for baseline details). This highlights our major motivation: unsupervised guidance can enhance report generation in scenarios lacking data labels.

\begin{table}[t]
\caption{Comparing report generation results from each clustering method on the thyroid dataset. B-1 to B-4 refer to BLEU-1 to BLEU-4. M and R-L denote METEOR and Rouge-L.}
\setlength{\tabcolsep}{4pt}
\begin{adjustbox}{width=\linewidth,center} 
\begin{tabular}{l | c c c c c c }
\hline
\textbf{Method} & \textbf{B-1} & \textbf{B-2} & \textbf{B-3} & \textbf{B-4} & \textbf{M} & \textbf{R-L} \\
\hline
\rowcolor{gray!20}
TF+SC$^1$ & 0.721 & 0.654 & 0.598 & 0.550 & 0.433 & 0.703 \\
\hline
DBSCAN \cite{ester1996density} & 0.728 & 0.663 & 0.608 & 0.561 & \textbf{0.501} & \textbf{0.726} \\
AC$^2$ \cite{mullner2011modern} & 0.718 & 0.659 & 0.607 & 0.564 & 0.484 & 0.722 \\
SC$^3$ \cite{von2007tutorial} & 0.717 & 0.656 & 0.603 & 0.558 & 0.487 & 0.697 \\
HDBSCAN \cite{mcinnes2017hdbscan} & 0.724 & \textbf{0.660} & 0.607 & 0.561 & 0.494 & 0.710 \\
K-Means \cite{ikotun2023k} & \textbf{0.729} &\textbf{ 0.666} & \textbf{0.613} & \textbf{0.568} & 0.439 & 0.723 \\
\hline
\multicolumn{7}{l}{$^1$ TF+SC means only adding the similarity comparison loss} \\
\multicolumn{7}{l}{$^2$ AC represents agglomerative clustering.} \\
\multicolumn{7}{l}{$^3$ SC represents spectral clustering.} \\
\hline
\end{tabular}
\end{adjustbox}
\label{table:cluster}
\end{table}

\subsection{Quantitative Results}
\label{sec:quantity}
To demonstrate the effectiveness of our approach, we compare our method with six other existing approaches:
\begin{itemize}
    \item CNN-RNN \cite{vinyals2015show}: This method first utilizes the CNN model to extract visual features from images and applies hierarchical LSTM decoding to generate reports.
    \item TriNet \cite{yang_joint_2021}: This method designs two branches to align visual and textual features. It is important to note that one branch in the original method requires medical subject headings, which are not available in our dataset. Therefore, this branch is removed from our comparison.
    \item R2Gen \cite{chen2020generating}: This method proposes a memory-driven unit to integrate memory into the Transformer, aiming to enhance the performance of radiology report generation.
    \item TF \cite{vaswani2017attention}: This method adopts the standard Transformer encoder-decoder framework. After extracting visual features from the image with CNN, these features are later inputted into the Transformer to generate text reports.
     \item R2GenRL \cite{qin2022reinforced}: This method is an improvement based on R2Gen. It enhances R2Gen with a reinforcement learning loss, using the BLEU-4 score as a reward to enhance report generation process.
     \item DeltaNet \cite{wu2022deltanet}: DetalNet is a retrieval-based report generation method. It retrieves the most similar medical images and reports from the training data based on the image, and employs them as references for the report generation.
\end{itemize}

Table \ref{table:comparison} represents the comparative results in the NLG and CE metrics. In the breast dataset, our method exhibits superior performance across most of the metrics. Compared to the second-ranking method (DeltaNet), the BLEU-1 to BLEU-4 metrics have increased by 6.3\%, 6.8\%, 5.33\%, and 5.26\% respectively. Similarly, in the thyroid dataset, our method demonstrates superior performance compared to other methods. Specifically, when compared to R2GenRL (the 2\textsuperscript{nd} best method), our approach shows a notable improvement of 20.74\% in accuracy. It is worth noting that the breast and thyroid datasets are larger compared to the liver dataset. Despite this disparity, our proposed method also demonstrates superior performance in the relatively smaller liver dataset, with the highest recall and F1 score, which means that as many key entities as possible are predicted in the reports.

\begin{figure*}[!t]
\centerline{\includegraphics[width=0.9\textwidth]{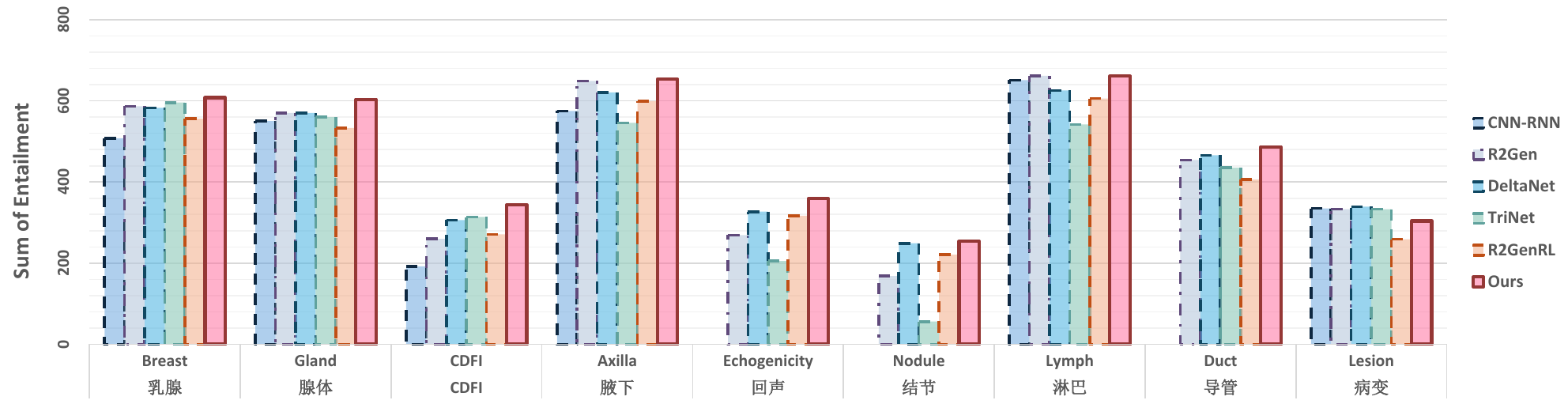}}
\caption{The number of correct entailments for different entities. A higher correct number represents more accurate descriptions per entity. It should be noted that the CNN-RNN method cannot describe certain entities, resulting in some entailment numbers of 0.}
\label{fig-entail}
\end{figure*}

In Table \ref{table:comparison}, we observe that our model achieves higher recall in all datasets. However, this does not necessarily imply an accurate prediction of the real meaning of the sentences. Therefore, in \mbluetext{Fig.} \ref{fig-entail}, we also assess the entailment between each key entity in the breast dataset. According to \mbluetext{Fig.} \ref{fig-entail}, the majority of the entities we predict align well with the original report and demonstrate the best performance. For example, regarding \say{echogenicity}, our method has the highest entailment number (359), while the second best is DeltaNet (327). Although in a few cases, such as \say{nodules}, we have similar performance compared to DeltaNet. Yet, in Table \ref{table:comparison}, we find that our method's parameter (60.251 M) is much smaller than DeltaNet (72.499 M). Hence, our method proves to be more suitable for real clinical settings, particularly in scenarios where computational resources may be limited.

\begin{table*}[ht]
\setlength{\tabcolsep}{5pt}
\caption{Performance Comparison from three ultrasound datasets. Best performances are highlighted in bold.}
\begin{adjustbox}{width=0.95\linewidth,center} 
\begin{tabular}{l | l | c | c c c c c c|c c c c}
    \hline
    \multicolumn{1}{l|}{\multirow{2}{*}{\textbf{Dataset}}} 
    & \multicolumn{1}{l|}{\multirow{2}{*}{\textbf{Method}}} 
    & \multicolumn{1}{l|}{{\textbf{Param.}}}
    &  \multicolumn{6}{c|}{\textbf{NLG METRICS $\uparrow$}}
    & \multicolumn{4}{c}{\textbf{CE METRICS $\uparrow$}}\\ 
    \multicolumn{1}{c|}{} 
    &\multicolumn{1}{c|}{} 
    &\textbf{(M)}
    & \textbf{BLEU-1} & \textbf{BLEU-2} & \textbf{BLEU-3} & \textbf{BLEU-4} & \textbf{METEOR} & \textbf{ROUGE-L}& \textbf{Accuracy} & \textbf{Precision} & \textbf{Recall} & \textbf{F1 Score} \\
    \hline
    \multirow{4}{*}{Breast$^1$} 
    & CNN-RNN \cite{vinyals2015show}& 7.189 & 0.114 & 0.093 & 0.078 & 0.067 & 0.221 & 0.185& 0.000 & 0.496 & 0.498 & 0.487 \\
    & TriNet \cite{yang_joint_2021}& 22.615  & 0.693 & 0.594 & 0.533 & 0.478 & 0.439 & 0.742& 0.351 & 0.816 & 0.697 & 0.727 \\  
    & R2Gen \cite{chen2020generating}&60.804   & 0.663 & 0.611 & 0.572 & 0.541 & 0.411 & 0.685& 0.494 & 0.800 & 0.761 & 0.776 \\
    & TF \cite{vaswani2017attention}& 60.232      & 0.699 & 0.653 & 0.619 & 0.590 & 0.437 & 0.757 & 0.461 & \textbf{0.827} & 0.671 & 0.702 \\
    & DeltaNet \cite{wu2022deltanet}&72.499      & 0.716 & 0.665 & 0.638 & 0.608 & \textbf{0.517} & \textbf{0.758} & 0.573 & 0.819 & 0.819 & 0.818 \\
    & R2GenRL \cite{qin2022reinforced}& 81.139      & 0.672 & 0.595 & 0.531 & 0.479 & 0.500 & 0.651& 0.424 & 0.793 & 0.754 & 0.771 \\
    \rowcolor{gray!25}& Ours& 60.251    & \textbf{0.761} & \textbf{0.710} & \textbf{0.672} & \textbf{0.640} & 0.468 & \textbf{0.758}  & \textbf{0.586} & 0.815 & \textbf{0.831} & \textbf{0.822}\\ 
    \hline
    \multirow{4}{*}{Thyroid$^2$} 
    & CNN-RNN \cite{vinyals2015show}& 7.189 & 0.131 & 0.105 & 0.086 & 0.069 & 0.069 & 0.207 & 0.000 & 0.448 & 0.348 & 0.382 \\
    & TriNet \cite{yang_joint_2021}& 22.615  & 0.645 & 0.510 & 0.421 & 0.345 & 0.409 & 0.678& 0.268 & 0.845 & 0.769 & 0.803 \\
    & R2Gen \cite{chen2020generating}&60.804   & 0.578 & 0.532 & 0.492 & 0.457 & 0.369 & 0.664& 0.404 & 0.810 & 0.768 & 0.779 \\
    & TF \cite{vaswani2017attention}& 60.232      & 0.709 & 0.642 & 0.585 & 0.538 & 0.425 & 0.701& 0.260 & 0.717 & 0.732 & 0.724 \\
 & DeltaNet \cite{wu2022deltanet}&72.499      & 0.610 & 0.559 & 0.515 & 0.579 & 0.443 & 0.685& 0.363 & 0.837 & 0.784 & 0.795 \\
    & R2GenRL \cite{qin2022reinforced}& 81.139      & 0.616 & 0.595 & 0.464 & 0.414 & \textbf{0.470} & 0.599 & 0.434 & 0.834 & 0.819 & 0.826 \\
    \rowcolor{gray!25}& Ours& 60.251    & \textbf{0.729} & \textbf{0.666} & \textbf{0.613} & \textbf{0.568} & 0.439 & \textbf{0.723}& \textbf{0.524} & \textbf{0.838} & \textbf{0.850} & \textbf{0.841} \\ 
    \hline
    \multirow{4}{*}{Liver$^3$}   
    & CNN-RNN \cite{vinyals2015show}& 7.189 & 0.049 & 0.026 & 0.011 & 0.000  & 0.119 & 0.102& 0.000 & 0.181 & 0.068 & 0.070 \\
    & TriNet \cite{yang_joint_2021}& 22.615  & 0.868 & 0.821 & 0.785 & 0.750 & 0.531 & 0.861& 0.039 & 0.898 & 0.809 & 0.814 \\
    & R2Gen \cite{chen2020generating}&60.804   & 0.866 & 0.842 & 0.822 & 0.805 & 0.537 & 0.869& 0.530 & 0.875 & 0.880 & 0.870 \\
    & TF \cite{vaswani2017attention}& 60.232      & 0.855 & 0.832 & 0.815 & 0.800 & 0.524 & 0.873& 0.444 & 0.749 & 0.785 & 0.765 \\
 & DeltaNet \cite{wu2022deltanet}&72.499      & \textbf{0.873} & 0.846 & 0.825 & 0.808 & 0.593 & 0.862 & \textbf{0.568} & \textbf{0.900} & 0.878 & 0.874 \\
    & R2GenRL \cite{qin2022reinforced}& 81.139      & 0.853 & 0.818 & 0.791 & 0.769 & 0.575 & 0.842& 0.466 & 0.885 & 0.875 & 0.879 \\
    \rowcolor{gray!25}& Ours& 60.251    & 0.872 & \textbf{0.848} & \textbf{0.828} & \textbf{0.813} & \textbf{0.539} & \textbf{0.875} & 0.541 & 0.879 & \textbf{0.894} & \textbf{0.883}\\ 
    \hline
\multicolumn{13}{l}{%
  \begin{tabular}[t]{@{}p{\linewidth}@{}}
    $^1$ In the breast dataset, the key entities include the breast, gland, Colour Doppler flow (CDFI), axilla, echogenicity, nodule, lymph node,  (mammary) duct, lesion, subcutaneous fat layer, and tumour. \\
    $^2$ For the thyroid dataset, the key entities are the thyroid gland, glandular tissue, echogenicity, lesion, CDFI, lymph node, border, shape, nodule, left lobe, right lobe, and margin (of the thyroid). \\
    $^3$ For the liver dataset, the key entities include liver, capsule, echogenicity, vein, kidney, intrahepatic duct, bile duct, gallbladder, margin (of the liver), pancreas, pancreatic duct, lesion, spleen, CDFI, and nodule.
  \end{tabular}%
 
} 
\\
\hline
\end{tabular}
\label{table:comparison}
\end{adjustbox}
\end{table*}

\begin{table}[t]
\caption{Ablation studies from three ultrasound datasets. Best performances are highlighted in bold.}
\setlength{\tabcolsep}{4pt}
\begin{adjustbox}{width=\linewidth,center}  
\begin{tabular}{l | l |c c c c c c}
    \hline
    \textbf{Dataset} & \textbf{Method} & \textbf{B-1} & \textbf{B-2} & \textbf{B-3} & \textbf{B-4} & \textbf{M} & \textbf{R-L} \\ 
    \hline
    \multirow{4}{*}{Breast} 
     & TF      & 0.699 & 0.653 & 0.619 & 0.590 & 0.437 & 0.757                      \\ 
     & TF+KMVE & 0.744 & 0.694 & 0.656 & 0.625 & 0.459 & 0.757                      \\ 
     & TF+SC   & 0.734 & 0.677 & 0.635 & 0.601 & 0.449 & 0.744                      \\ 
     & Ours    & \textbf{0.761} & \textbf{0.710} & \textbf{0.672} & \textbf{0.640} & \textbf{0.468} & \textbf{0.758} \\ 
     \hline
    \multirow{4}{*}{Thyroid} 
     & TF      & 0.709 & 0.642 & 0.585 & 0.538 & 0.425 & 0.701                      \\ 
     & TF+KMVE & 0.719 & 0.658 & 0.607 & 0.564 & 0.436 & 0.723                      \\ 
     & TF+SC   & 0.721 & 0.654 & 0.598 & 0.550 & 0.433 & 0.703                      \\ 
     & Ours    & \textbf{0.729} & \textbf{0.666} & \textbf{0.613} & \textbf{0.568} & 0.439 & \textbf{0.723} \\ 
     \hline
    \multirow{4}{*}{Liver} 
     & TF      & 0.855 & 0.832 & 0.815 & 0.800 & 0.524 & 0.873                      \\ 
     & TF+KMVE & 0.857 & 0.835 & 0.817 & 0.802 & 0.525 & 0.875                      \\ 
     & TF+SC   & 0.856 & 0.834 & 0.817 & 0.802 & 0.524 & 0.875                      \\ 
     & Ours & \textbf{0.872} & \textbf{0.848} & \textbf{0.828} & \textbf{0.813} & \textbf{0.539} & \textbf{0.875} \\ 
     \hline

\end{tabular}
\end{adjustbox}
\label{table:ablation}
\end{table}

\begin{figure}[h]
\centerline{\includegraphics[width=0.8\columnwidth]{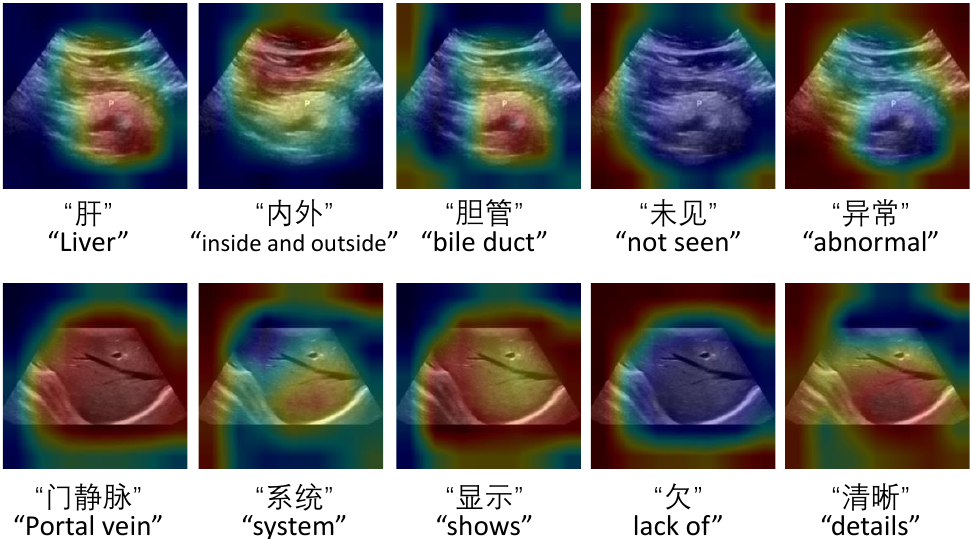}} 
\caption{Attention maps of different words in the sentence. Warm colours indicate high attention, while cool colours indicate low attention. To preserve the original word order of Chinese, the English translation may exhibit unconventional expressions and grammatical variations.}
\label{fig:attn}
\end{figure}

\subsection{Ablation Study}
\label{subsec:ablation}

In this section, we conduct ablation experiments to verify the effectiveness of each module, and the experimental results are shown in Table \ref{table:ablation}. The experiments include results for the following configurations: TF (using only the Transformer model), TF+KMVE (adding KMVE loss), and TF+SC (adding SC loss). Our proposed method represents a complete framework combining SC and KMVE losses. 

Analyzing the results in Table \ref{table:ablation}, we observe an  improvement when incorporating the KMVE loss during training. Specifically, in the breast dataset, we achieved the highest improvement in BLEU-1 with a 4.5\% increase. In the thyroid dataset, the most notable improvement is observed in BLEU-4, with an increase of 2.6\%. Additionally, in the liver dataset, we observe a slight increase across various metrics, with BLEU-2 increasing by 0.3\%. These results indicate that the KMVE module contributes to generating more accurate ultrasound reports, particularly in terms of n-gram matching and overall sentence quality. Furthermore, when adding the SC loss by incorporating the SC module into the framework, we also observe an increase in most metrics across the three datasets. However, the proposed module exhibits less improvement in the liver dataset compared to the breast and thyroid datasets. This might be the smaller size of the liver dataset and the basic method already achieves a high BLEU-4 score (0.80), indicating a strong similarity between generated and ground-truth results, leaving less room for improvement.

In summary, the experimental results validate the effectiveness of our proposed framework. The incorporation of KMVE modules enhances text generation quality, particularly in terms of n-gram matching and overall sentence quality. Additionally, the SC module provides further performance improvements by evaluating semantic consistency. 

\begin{figure*}[t]
\centerline{\includegraphics[width=0.95\textwidth]{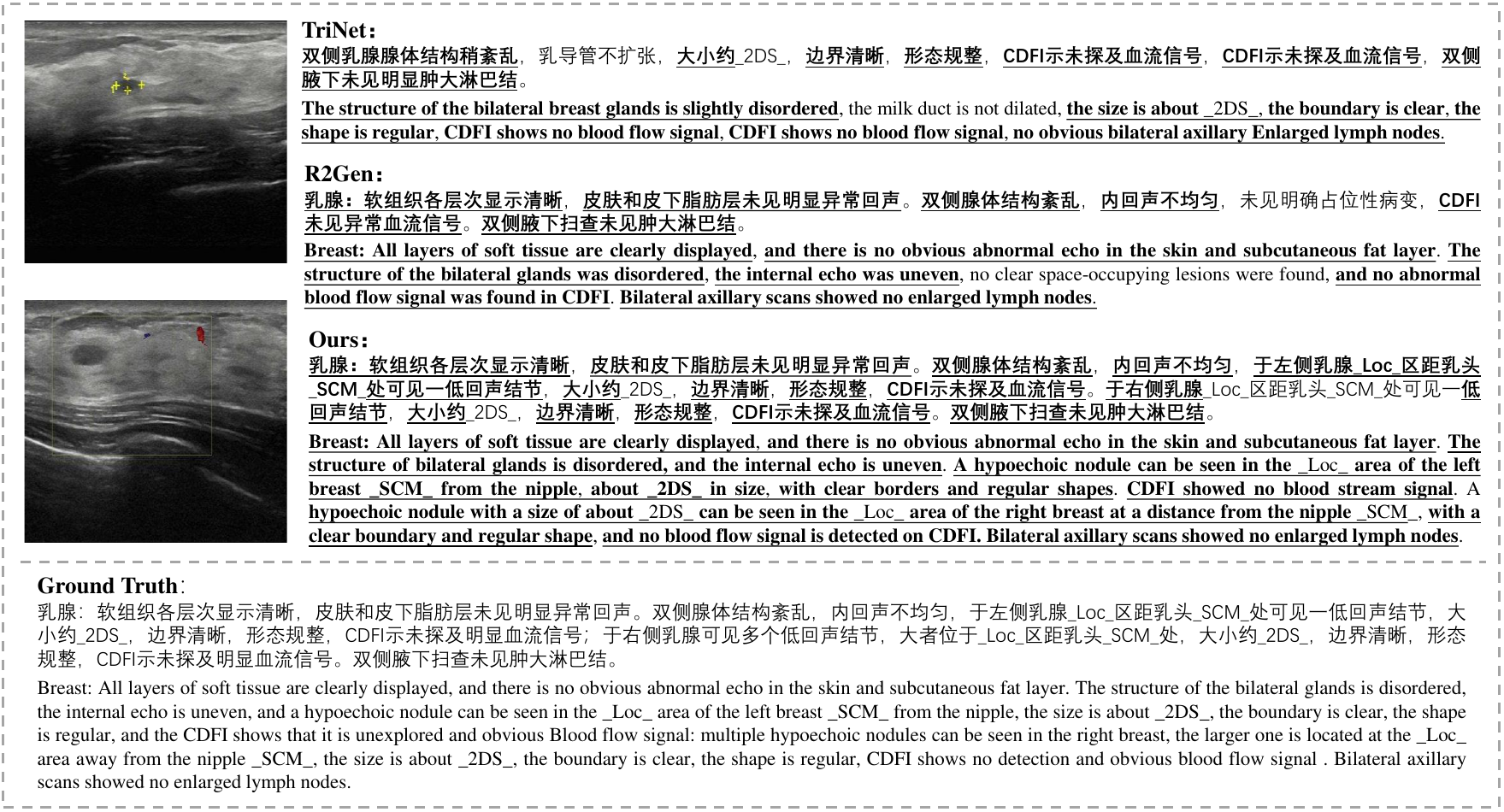}}
\caption{Visualization Results in Breast Dataset. The highlighted words represent descriptions aligned with the ground truth reports. It is worth noting that all the results were originally presented in Chinese. We used Google Translate to provide the English translations.}
\label{fig5}
\end{figure*}

\begin{figure*}[ht]
\centerline{\includegraphics[width=0.95\textwidth]{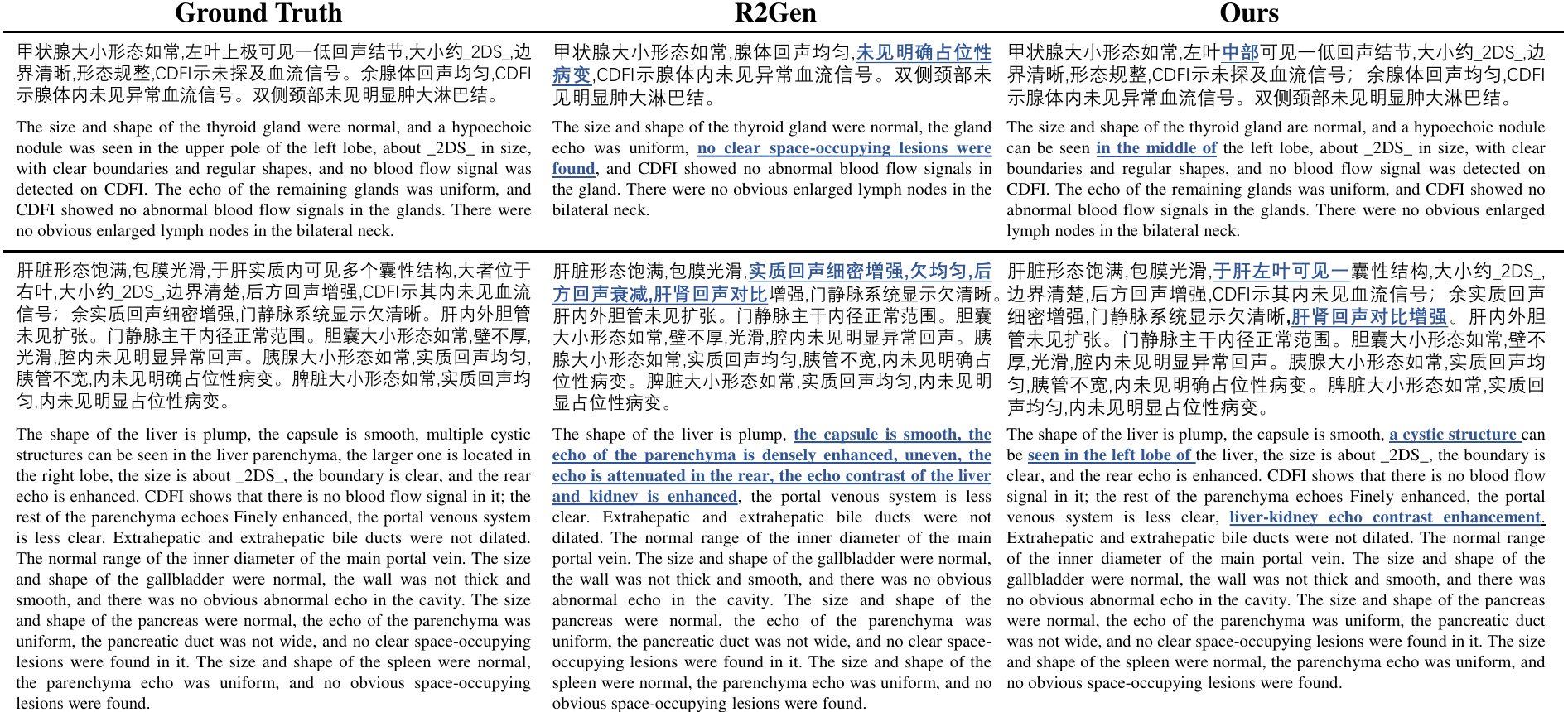}}
\caption{Visualization Results in Thyroid and Liver Dataset. The highlighted words are incorrect and differ from the ground truth reports. It is worth noting that all the results were originally presented in Chinese. We used Google Translate to provide the English translations.}
\label{fig6}
\end{figure*}

\subsection{Visualization Results}
\label{sec:visulize}
\mbluetext{Fig.~\ref{fig5}} presents the outcomes of the ultrasound report generation on the breast dataset, with the bold underlined sentences indicating semantically equivalent statements to the ground truth reports. The results highlight our method's better ability to generate crucial details compared to other approaches. Our approach successfully achieves a balanced representation of normal and abnormal descriptions, closely resembling the ground truth. For instance, when considering normal descriptions such as soft tissue, skin, and subcutaneous fat layer, the TriNet method fails to generate that, while our method and the R2Gen method accurately describe these features. For the crucial pathology of hypoechoic nodules, our method identifies the presence of lesions in the given images, providing a precise description: \say{A hypoechoic nodule was seen in the \texttt{\_Loc\_} area of the left breast \texttt{\_SCM\_} from the nipple.} This aligns with the sentences in the ground truth. In contrast, TriNet and R2Gen fail to capture this crucial finding. Besides, our method excels in imitating the writing style of real reports, including both length and sentence structure. \mbluetext{Fig.~\ref{fig6}} illustrates the results obtained on the thyroid and liver datasets. For the thyroid dataset, both R2Gen and our method offer accurate descriptions of the thyroid, including the CDFI blood flow signal and the bilateral neck. However, R2Gen fails to capture the essential description of abnormal nodules, which our method successfully includes. Regarding the liver dataset, both our method and R2Gen achieved satisfactory results in the generated reports, with a minimal difference between them. By analyzing the results, we observe that our proposed method effectively includes both normal and abnormal descriptions in the generated ultrasound reports, resulting in reports that closely resemble those written by doctors in clinical settings. 

Nevertheless, our method still encounters certain challenges, particularly in the fine-grained aspects of the reports. For instance, in \mbluetext{Fig.~\ref{fig6}}, our method describes the location of thyroid hypoechoic nodules as “the middle of the left lobe", while the accurate description should be \say{the upper pole of the left lobe}. When referring to cystic structures in the liver, the true report states \say{multiple cystic structures can be seen in the liver}, while our method generates \say{a cystic structure can be seen in the left lobe of the liver}. These examples indicate that our method may be less sensitive to accurately determining the number and precise location of lesions. The aforementioned challenges are not unique to our method but are common issues faced by the current field of ultrasound report generation. Unlike publicly available datasets for lung X-rays, such as IU-Xray and MIMIC-CXR, the natural variations in ultrasound reports are more diverse. Ultrasound is utilized for disease screening in various organs, enabling accurate descriptions of specific lesions, including details such as location and size. Thus, we firmly believe that further research and discussion are crucial in exploring this direction.

Furthermore, we visualize the attention map of our model at the word level in Figure \ref{fig:attn}. It reveals that our model allocates varying degrees of attention to each word. Notably, essential terms like \say{liver} and \say{bile duct} receive focused attention, aiming to pinpoint their respective locations within the image. On the other hand, terms like \say{lack of}, \say{not seen} receive comparatively less attention. This is likely because of their abstract nature, which is not easily represented visually. These attention maps offer valuable insights into how our model processes information at each word.

\section{Conclusion}
In this work, we propose a novel framework that combines unsupervised and supervised learning for ultrasound report generation. Our framework leverages the unsupervised learning clustering method to extract prior knowledge from ultrasound text reports, which is then utilized to guide the training process. Additionally, we designed a similarity comparer in the report generator to enhance the prediction process. Furthermore, we built three large ultrasound report datasets of different organs to assess the framework's performance across various organs. Through extensive experimentation and analysis of three ultrasound datasets, we demonstrate the effectiveness and superiority of our framework compared to baseline models. Despite the promising results achieved, it is important to realise the existing limitations of current models. Similar to other state-of-the-art approaches, our method exhibits insensitivity to terms related to size, location, and number within ultrasound reports. This insensitivity may be attributed to the uneven distribution of these terms in the vocabulary, following a long-tail distribution pattern. Consequently, further research is required to address this challenge and improve the accuracy in handling such situations.

\bibliographystyle{ieeetr}
\bibliography{libarary.bib} 

\end{document}